\documentclass{article}

\usepackage[preprint]{neurips_2021}

\usepackage[utf8]{inputenc} 
\usepackage[T1]{fontenc}    
\usepackage{hyperref}       
\usepackage{url}            
\usepackage{booktabs}       
\usepackage{amsfonts}       
\usepackage{nicefrac}       
\usepackage{microtype}      
\usepackage{xcolor}         
\usepackage{graphicx}       

\title{Doing Natural Language Processing in A Natural Way:
An NLP toolkit based on object-oriented knowledge base and multi-level grammar base}

\author{%
  Yu Guo \\
  China Mobile Research Institute\\
  \texttt{guoyu0227@pku.edu.cn} \\
}

\begin{document}

\maketitle

\begin{abstract}

We introduce an NLP toolkit based on object-oriented knowledge base and multi-level grammar base. This toolkit focuses on semantic parsing, it also has abilities to discover new knowledge and grammar automatically, new discovered knowledge and grammar will be identified by human, and will be used to update the knowledge base and grammar base. This process can be iterated many times to improve the toolkit continuously.

\end{abstract}

\section{Introduction}

The main function of language is to convey and record information. People extract information from language and convert information in their mind into language. If we wish a machine to understand human language, we want the machine to be able to understand information, we already have RDF (Resource Description Framework) and OWL (Web Ontology Language) to process information for machine learning, but we think they are not powerful enough to describe the real world, so we develop an object-oriented Information Description Framework (IDF) inspired by the object-oriented programming language. In such framework, everything in the real world is considered as an object, language is considered as something describe various behaviors of the objects and the value of a certain property of an object. 

However, not all information is meaningful, only verified information should be recorded and stored, which is the so-called "knowledge". For a long time, people have been trying to introduce knowledge into natural language processing (NLP) tools. Many general-purpose knowledge graph are built for this goal, for Chinese, such as Wki(ZH)\cite{wiki}, CN\_Pedia\cite{DBLP:conf/cikm/ZhangXTW014}, Pkubase\cite{8509486}, Xlore\cite{4633358}, etc. We merged these knowledge graphs together into our own knowledge base according to our object-oriented IDF.

NLP tools based on grammar rules have better explainability, they are more similar to the way how human understand language. However, the number of grammar rules is huge, the linguistic definition of some special usages is ambiguous, and it is difficult to parse complex sentences containing a variety of grammar rules. All these problems make the performance of these tools worse than the statistical language models and deep learning language models.

We do not intend to study linguistics, so we designed a multi-level grammar description framework (GDF) to describe all kinds of grammars in natural language, without going deep into the grammatical principles behind the grammars. This framework also describes how to extract information from a certain grammar.

Based on the object-oriented IDF and multi-level GDF, we developed an NLP toolkit to build knowledge base and grammar base, parse natural language, and extract information. We call this toolkit "Sosweety". The code of this toolkit has been released in my github\footnote{https://github.com/ss433s/sosweety}.

The knowledge base has a lot of mistakes and some knowledges are missing, the grammar base need to be built from scratch. Also,language is not static, new words and new grammars are being created all the time. So, we add several learning abilities to discover knowledges and grammars automatically in our toolkit, however, in this very early version, the final determination whether the new discovered knowledge or grammar can be accepted will be made by humans.

The whole working flow of this toolkit are showed in Figure 1.

\begin{figure}
  \centering
  \includegraphics[scale=0.3]{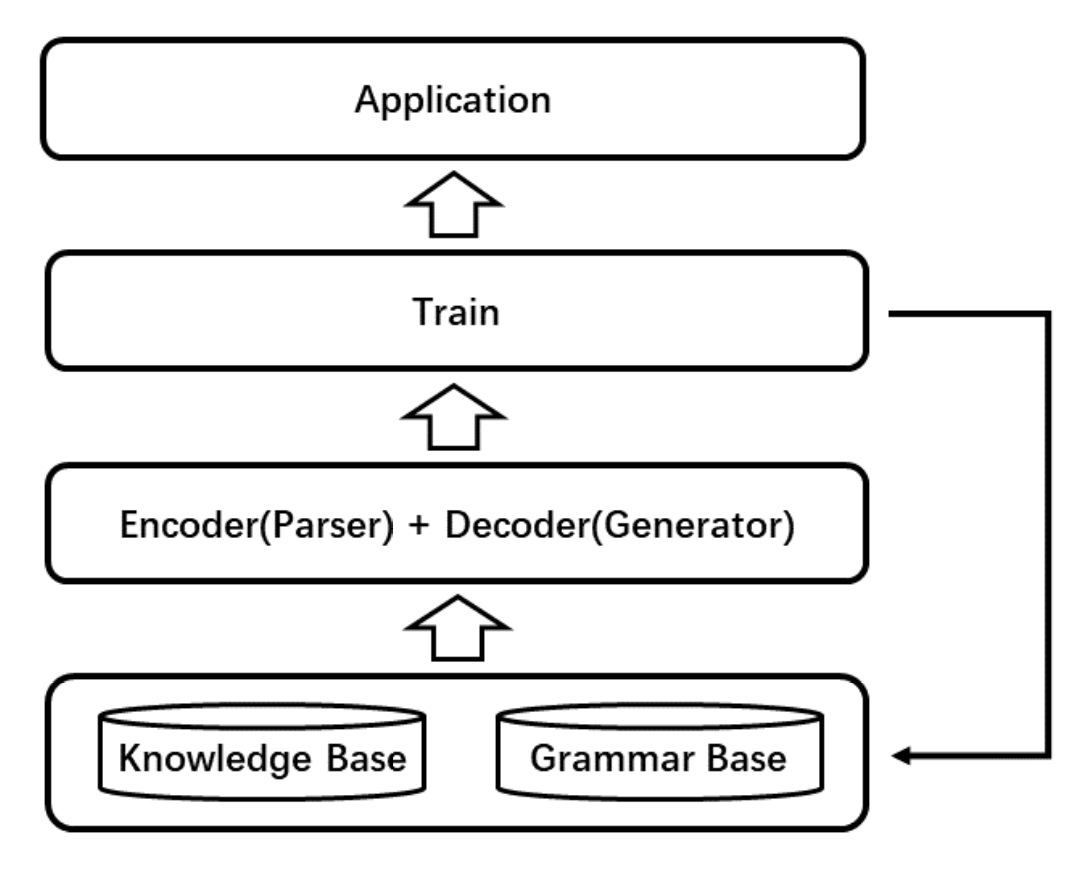}
  \caption{Working flow of the toolkit.}
\end{figure}

So far, the knowledge base has approximately 11 million entities, 40,000 methods, and 24 million relations. The grammar base has 81 phrase patterns and 178 subsentence patterns. All the experiments in this paper are performed by a Thinkpad X390 notebook with Intel i5 processor, 8GB memory, 512GB SSD. Training with a corpus which contains 170,000 sentences could be done within 2 hours in fast mode.

The contributions of our paper are as follows:

1, We design a set of object-oriented Information Description Framework.

2, We design a set of multi-level Grammar Description Framework.

3, We develop an NLP toolkit to build knowledge base and grammar base, parse natural language, and extract information from natural language.

4, We achieve some learning abilities for unknown knowledge and grammar semi-automatically.

\section{Object-oriented Information Description Framework}

\subsection{Object, Property, Method}

In this framework, everything is an object, objects have properties and methods, just like other object-oriented programming language. Unlike most of the other object-oriented programming languages, properties and methods are also objects in our framework. For example, "Xiao Ming" is an object, he has properties such as "name", "name" is also an object, it has properties such as "length".

\subsection{No class and instance}

Another difference is that, in this framework, there is no class and instance, class and instance are merged as "Concept". In many cases, it is difficult to distinguish class and instance. For example, "Xiao Ming" is an instance of the "Person" class, but when we want to mention "18-year-old Xiao Ming", "Xiao Ming ten years ago", "future Xiao Ming", it seems that "Xiao Ming" can also be treated as a class, "Xiao Ming" at different moments are instances of the "Xiao Ming" class. For another example, in the classification of species, we have kingdom, phyla, family, genus and species. We take "Species" as a class, all kinds of species are instance of this class. Then, "Person" or "Human", which we take as a class in the above example, should be an instance of "Species".

Therefore, we decide to remove class and instance, and unify them as "Concept". Each concept has its own properties, methods, and sub-concepts. Concept inherits the properties and methods from its parent concepts, however, not all the properties and methods will be inherited. For example, concept "Human" has method "Look", the sub-concept "Blind person" cannot inherit this method.

It is difficult to program with such paradigm, but we think it is necessary for describing the real world.

\subsection{Relationship between concept and method}

In the real world, which properties a certain concept has, which methods can be implemented by a certain concept, and which concepts a method can act on are also part of human knowledge. These knowledges should not be ignored in an information description framework, and it will have a great benefit in grammar analysis, which can help eliminate many impossible combinations between concepts and methods.

For example, "Person" is a concept, "Person" has properties such as "name", "gender", "Person" can implement methods such as "eat" and "sleep", "Person" cannot implement "thunder" or "rain". "Play" is a method, and the target of this method can be a "Ball", but not a "Table". We can even write a piece of code for some methods and execute the code when needed.

We use two tables "Concept\_tbl" and "Method\_tbl" to record the concept ids and method ids and their relationships. For concept, we have a "properties" field to record which properties this concept has, a "methods" field to record which methods this concept can implement. For method, we have an "objects" field to record which concepts a method can act on, a "code" field to record a piece of code of how this method be implemented.

\subsection{Behavior and property}

The value of properties of a concept and the behavior of a concept are stored in the "Fact\_tbl" just like triples in RDF.

\subsection{Logical Relationship}

There are several kinds of logical relationships in the world, we take the causal relationship as the most important one. In our knowledge base, causal relationships are recorded between "facts", "facts" of the same method
 will be merged. For example, "I hit Jim" is the reason of "Jim get hurt", "Sam hit Lily" is the reason of "Lily get hurt", then we can merge all this type of "facts" into one logical relationship that "hit" can cause "get hurt".

As our toolkit is still at the stage of semantic parsing, the fact and logical relationship part will be completed in the future.

\subsection{Word linking}

In natural languages, one word can have multiple meanings and one concept can be expressed by multiple words. For verbs and adjectives, similar phenomena are also common, such as "see" and "look". 

So only entity linking is not enough, in our knowledge base, we use a "Word\_tbl" to record the many-to-many relationship between words and objects (concept, method, etc.).

This is the only interface between the language and the knowledge base. All text will be converted into object ids through this table, and other tables in the knowledge base will only store object ids. All other calculations will be based on object ids.

\section{Multi-level Grammar Description Framework}

There are many customary rules in human language. It is difficult to trace the origin of these rules, and some of them are difficult to define in linguistics. Unlike linguists, we do not try to analyze the linguistic principles for various grammar rules. The grammar base just records the rules in a way which a machine can recognize.

For example, "I will send her a bunch of flowers", is the object of "send" "her" or "flower"? For Chinese, linguists have different opinions, but we only need to know that there is a structure such as "noun1|verb|noun2|noun3" to indicate that noun1 implement "verb1" to "noun2", the object of the action is "noun3". In fact, not all verbs can be used in this structure. We also need to record which verbs can be used here.

There are different levels of structures in our language, in Chinese, we have characters, words, phrases, subsentences, sentences, paragraphs, articles, books, etc. Each level has its own patterns, therefore, our grammar description framework records the patterns separately for different levels.

At this stage, we mainly focus on the phrase patterns and subsentence patterns.

\paragraph{Character}

The Chinese characters have some special components, these components can help understanding the meaning of the character. The character patterns can help recognize a new character for humans, but maybe not very useful for a machine, as we can load all the characters into the grammar base.

\paragraph{Word}

One word can have multiple meanings and one concept can be expressed by multiple words. Some words sharing the same characters has similar meanings. But we don't have the word level patterns, instead, we created a new level "Concept".

\paragraph{Concept}

Unlike English, Chinese does not use space between words, it is hard to distinguish word from phrase in some cases. 

So, we added an intermediate level "Concept", which corresponds to the "Concept" in the knowledge base. No matter a string is considered as a "Word" or "Phrase", if it is a "Concept" in the knowledge base, we take it as "Concept".

\paragraph{Phrase}

If we only consider the part-of-speech (POS), there won't be too many phrase patterns, but if we want to extract information from the phrase correctly, we will need much more patterns. It is the most complex part of the grammar base.

\paragraph{Subsentence}

In our grammar description framework, we define the part of a sentence separated by punctuation marks such as commas and semicolons as subsentence. Some sentences contain only one subsentence, and some contain several subsentences.

Some subsentences, especially in question and answer, have only one word or one phrase, we define this type of subsentence as a "phrase" type subsentence; some have a complete semantic structure, such as the most common " subject- predicate-object" structure in both Chinese and English, we define this type of subsentence as a "sentence" type subsentence; some subsentences lack a certain semantic component, we define this type of subsentence as a "half\_sentence" type subsentence, the missing part of this kind of subsentence is often inherited from other subsentences or other sentences.

Similar to English, Chinese sentences can also be divided into four categories: declarative sentences, interrogative sentences, imperative sentences, and exclamatory sentences. Different type of sentences may have the same structure, but have different meaning, so we add another type to record this difference.

\paragraph{Sentence}

Some sentences contain only one subsentence, and the rules for these sentences are consistent with subsentence. 

Some sentences combine various subsentences together with conjunctions. The pattern of this kind of sentences are also recorded. These conjunctions, such as "because" or "as", often represent a certain logical relationship, these relationships will be part of the information extracted from the sentences. 

Some sentences contain multiple subsentences, but there are only simple inheritance relationships between the subsentences. For these sentences, there are no patterns, we will parse each subsentence separately, for "half\_sentcence" subsentences, inheritance source of the missing part will be identified.

The sentence level pattern will not be released in this version.

\paragraph{Paragraphs, chapters, books, etc.}

These types of text have their own patterns. Currently, we still focus on parsing for sentences, these higher levels will be considered in the future.

\section{Grammar Base}

There are only tens of thousands of Chinese characters and hundreds of thousands of basic words, all of which can be recorded through the vocabulary of the knowledge base. Complex sentences and higher-level text structures will be considered in the future. Right now, the grammar base mainly consists of phrase patterns and subsentence patterns. The POS and dependency relation symbols are following the rule of Stanfordcorenlp\cite{manning-EtAl:2014:P14-5} with some modification.

\subsection{Process phrase just like word}

As mentioned above, phrase patterns contain the rules for both concepts and phrases. 

In order to maintain the consistency of words and phrases in the parsing process, we encapsulate words and phrases to two classes: "Word" and "Phrase". Both "Word" and "Phrase" have attributes of "value", "pos", "core\_word", where "value" is the character string of the word or phrase; "pos" is the POS, for Word, it is the POS of the word itself, for Phrase, it is defined by the "pos\_tag" field of the phrase pattern; "core\_word" is the main word of a phrase, this attribute is used to determine the subject-predicate-object relationship and concept relationship. For example, "a beautiful car", the core word is "car". If this phrase is used as a subject, then the predicate it receives should be the method that the concept of "car" can implement. The upper-level concept of this phrase is also the upper-level concept of "car", such as "vehicle". For Word, "core\_word" is the word itself, for Phrase, there is a "core\_word\_index" field in the phrase pattern, this field is used to indicate which element in the phrase is the core word, the "core\_word" of the entire phrase is the "core\_word" of this element. This element can be either Word or Phrase, which means "core\_word" is recursive. If "core\_word\_index" is empty, "core\_word" is equal to "value". 

Following this way, both words and phrases can be processed equally. So, when setting the POS of the phrase, we do not use additional symbols such as "NP" and "VP", but still used symbols such as "NN", "VV" just like the POS of word.

\subsection{Phrase pattern}

A phrase pattern has "core\_word\_idnex", "pos", "meaning", "features" and some other attributes. 

"Core\_word\_idnex" is the index of the core word of a phrase, "pos" is the POS of a phrase."Meaning" is the semantic relationships in the phrase.

"Features" is a list of "feature". "Feature" is the basic element of a phrase pattern. "Feature" has three types: "word", "concept" and "pos". The value of a "word" feature is a word, only this word can match this feature. The value of a "concept" feature is a concept id. For example, we have a "concept" feature, its value is "A", if we identify a word is a concept "B" in the knowledge base and the concept "B" is a sub-concept of "A", this word will be considered to match this feature. The value of a "pos" feature is a POS, every word of this POS is considered to match this feature.

We use a tupple to represent a feature, for example, ("word", "basketball") means this is a "word" type feature, its value is "basketball". For a phrase pattern [("word", "basketball"), ("word", "player")], which contains two "word" features,  only the phrase "basketball player" can match this pattern. For a phrase pattern [("concept", 332), ("word", "player")], the concept whose id is 332 is "sport", concepts, such as "football" or "volleyball", which belong to "sport", can match this feature, so both "football player" and "volleyball player" can match this pattern.

\subsection{Subsentence Pattern}

A subsentence pattern has "parse\_str", "ss\_type", "ss\_type2", "meaning" and some other attributes.

A phrase that matches a phrase pattern will be merged into a new element, replaces the words or phrases belong to this phrase in the original parse result. When all the phrases in a subsentence have been correctly found and replaced, what remains is the main structure of the subsentence, the POS of all the elements in the main structure will be joint together with a delimiter "|" to form a "parse\_str". This "parse\_str" is the unique identifier of a subsentence pattern. 

There are three types of "ss\_type": "sentence", "half\_sentence", and "phrase", which respectively represent complete subsentences, subsentences with missing components and subsentences that contain only one phrase.

There are four types of "ss\_type2": "d", "q", "i" and "e", which respectively represent declarative sentences, interrogative sentences, imperative sentences, and exclamatory sentences.

The "meaning" field records the dependency relationships between the components of the subsentence. At present, only subject-predicate relationship and verb-object relationship are supported. A subsentence can have multiple relationships, each relationship is separated by a comma. The first field in the relationship is the relationship type, for subject-predicate relationship is "nsubj", for verb-object relationship is "dobj". The second and third fields are indexes of the head and tail entity of the relationship. The fields in a relationship are separated by colons.

For example, a subsentence pattern whose "parse\_str" is "NN|VV|NN", "ss\_type" is "sentence", "ss\_type2" is "d" and "meaning" is "nsubj:0:1,dobj:1:2". This is the most common subsentence pattern of the subject-predicate-object structure in both Chinese and English. The type of subsentences matching this pattern is a complete sentence. The first "NN" and "VV" form a subject-predicate relationship, "VV" and the second "NN" form a verb-object relationship.

\section{Encoder and Decoder}

The final aim of this toolkit is to achieve the conversion between language and information. We define the process of information extraction from language as encode, and generation language from information as decode.

\subsection{Encoder}

Someone has tried to extract information based on dependency parse. The problem is it is hard to get the right dependency relationships, and some dependency relationships are not detailed enough to extract the exact information.

With the help of the knowledge base and grammar base, we can achieve the dependency parsing more correctly. We add the "meaning" attribute to all the phrase patterns and subsentence patterns, make it more detailed to extract information from the sentences.

For simple sentences, we can traverse all possible combinations for all grammars. For complex sentences, this process is too time-consuming So, we designed two algorithms to accelerate this process:

1, Single recursion. Loop through all phrase patterns, and if there is a match, replace the matched elements to a new phrase and jump out of the loop. Repeat this loop until no matching phrase pattern can be found.

For complex sentences, it is difficult for this algorithm to find the correct structure, but it is the fastest way for simple sentences. This algorithm is mainly used in the process of initialization and fast learning.

2, Attention. Like human, at the first sight, this algorithm will focus on the most common words in the sentence to guess the main information of the sentence, and then complete the full information extraction with the other components. There are still some problems in the algorithm to be solved.

\subsection{Decoder}

The Decoder part will be completed in the future.

\section{Learning Ability}

There are so many knowledges and grammars, it is impossible to load all of them manually. Every year, new vocabulary and grammar are created. Luckily, it is easy to add new knowledges and grammars to our knowledge base and grammar base. The only problem is how to find them.

In our toolkit, we have built several learning abilities to discover new knowledges and grammars automatically, but the final determination will be done manually at this stage.

\subsection{Concept Formulation Rules Discovery}

Words belongs to a certain concept often has some rules in Chinese. For example, we know that "Shandong Province" and "Shanxi Province" belong to the concept "Province", then we can guess that another "xx province", for example "Zhejiang Province", is a "province" too.

In this module, now we have only one algorithm. We check the first 1-2 characters or the last 1-2 characters for all the words belong to a certain concept, if all the sub-concepts have the same characters at the same position, we take this as a potential rule. If the proportion of sub-concepts in a certain concept conforming this rule exceeds a certain threshold, and the proportion of the words belonging to this concept in all words conforming to this rule exceeds a certain threshold, this rule will be identified as a valid one.

\subsection{New Concept Discovery}

Like n-gram, if some words often form a fixed combination in the sentences which cannot be parsed, the algorithm will take this combination as a potential concept.

\subsection{New "Concept" Feature Discovery}

Extract high frequency words from sentences that cannot be parsed, and define all the words that have appeared before or after the high-frequency words as a new concept. For example, we have three phrases, "love movie", "comedy movie", "costume movie", we find that these three words, "love", "comedy" and "costume", form a phrase at the same position with the same word "movie", then the module will take these words as the sub-concepts of the same parent concept.

\subsection{New Phrase Patterns Discovery}

Here we present two ways to do this. First, check out all Tri-Gram words combination, then, checkout all the possible upper concepts for all the words, traverse all possible combinations to figure out which combinations appears most often in the corpus. This method requires too much calculation so that we only check out the combinations which contain only nouns. Second, extract high frequency words from sentences that cannot be parsed, check if there is another word that appears together with this high frequency word around, check out all combinations between the two words to find out if there is a potential phrase pattern.

\subsection{New Subsentence Patterns Discovery}

Find new subsentence patterns from sentences cannot be parsed. Check out phrases in the sentence as much as possible, take what remains as new subsentence patterns.

\section{Experiment}

All the experiments are performed in the WSL environment on a Thinkpad X390 notebook configured with an Intel i5 processor, 8GB memory, and 512GB SSD.

In our experiments, we find that phrase patterns containing "concept" features is the most numerous of all rules. When performing new subsentence pattern discovery, we find that most new patterns we find are not truly subsentence pattern, but patterns with an unrecognized "concept" features. Most are caused by the mistakes and deficiencies of the knowledge base. So, at present, we are still trying to identify this type of phrase patterns as many as possible, but clean of the knowledge base will be necessary.

\subsection{Initialize Knowledge Base}

The main source of the knowledge base is the general-purpose knowledge graphs, PKUBASE, CN\_Pedia and another graph we built based on Wiki(ZH).

The knowledge base is a sqlite database, which has 5 tables, such as Concept\_tbl, Method\_tbl, Fact\_tbl, Word\_tbl, Concept\_relation\_tbl.
Concept\_tbl has ~11 million concepts, Method\_tbl has ~40,000 methods.

The relations between concepts and methods were identified in another project, in which we used three NLP tools, Stanfordcorenlp, HanLP and LTP, parsed ~10 million sentences of Chinese wiki corpus, the dependency parse result was transferred to the relation between concepts and methods. Concept\_relation\_tbl has ~24 million relations. Now there is only one type of relations, which is "A belong to B".

Fact\_tbl is empty at this stage.

The whole knowledge base can be constructed within 10 minutes, the final database file is ~3GB. Here is a trick for use, we copy this file to /dev/shm, make the database run in memory, this will improve the speed significantly. (This trick doesn’t work for WSL1)

\subsection{Initialize Grammar Base}

The corpus is from the Information Extraction task of the 2019 Language and Intelligence Challenge, which contains ~170,000 sentences.

We pack the learning abilities into a learning module, and use this module to analyze the corpus, the raw semantic parse is performed by Standfordcorenlp, not by our own parser.

55 phrase patterns and 18 subsentence patterns are selected to initialize the grammar base.

\subsection{Iterate Grammar Base }

With the initialized grammar base, the parsing and learning pipeline are performed to the corpus again.

The parsing is done with the single recursion algorithm, about 40\% sentences can be parsed. The rest are sent to the learning pipeline, 18 new phrase patterns and 45 subsentence patterns are found, some of them are selected to be added to the grammar base.

\subsection{Question Patterns Analysis}

Most of the sentences of the information extraction corpus are declarative sentences, so we add another corpus from the Machine Reading Comprehension (MRC) task of the 2019 Language and Intelligence Challenge. This corpus contains ~400,000 questions and answers, we only use the questions for our experiment. We add a type of word "question word" which has a pos tag "QA", and try to figure out how these words works in Chinese.

8 phrase patterns and 115 subsentence patterns are identified and added to the grammar base.

\section{Future Work}

This toolkit is still a very early version, and many functions have not yet been realized. The current plans are:

1. Improve the parser. It is planned to add 1) POS correction, 2) construction of sentence pattern library, and 3) mistakes recognition and correction. The final aim is that the toolkit can perform sentence parsing and information extraction to any text.

2. Discovery of logical relations.

3. Knowledge base clean.

4. Decoder construction. Enable the toolkit to convert the information in the knowledge base into natural language.

5. Clear the knowledge base and grammar base, and try to summarize the knowledge and grammar completely by the program from the large scale corpus without manual annotation.

\section{Discussion}

Unlike neural networks which imitate human neurons, we have created a system that imitates the logical thinking of humans.

At present, it has not been able to initialize and update the grammar base and knowledge base automatically, the algorithm of parsing is not perfect, and the effect of parsing is not as good as the effect of neural network language model.

However, all the procedure of the parsing process are made according to the grammar rules and knowledge, rather than billions of parameters in a neural network, which makes it easy to be explained to humans. Also, the toolkit can discover new grammar and knowledge by itself, and the newly discovered grammar and knowledge can be easily updated and operated, which makes the system easier to evolve.

We hope that more people can try to use this way for natural language processing, and improve the toolkit together.

\bibliographystyle{unsrt}

\begin{thebibliography}{1}

\bibitem[Wikipedia(2004)]{wiki}
Wikipedia.
\newblock Plagiarism --- {W}ikipedia{,} the free encyclopedia, 2004.
\newblock [Online; accessed 22-July-2004].

\bibitem[Zhang et al.(2014)]{DBLP:conf/cikm/ZhangXTW014}
Kezun Zhang, Yanghua Xiao, Hanghang Tong, Haixun Wang, and Wei Wang.
\newblock Wiicluster: a platform for wikipedia infobox generation.
\newblock In Jianzhong Li, Xiaoyang~Sean Wang, Minos~N. Garofalakis, Ian
  Soboroff, Torsten Suel, and Min Wang, editors, {\em Proceedings of the 23rd
  {ACM} International Conference on Conference on Information and Knowledge
  Management, {CIKM} 2014, Shanghai, China, November 3-7, 2014}, pages
  2033--2035. {ACM}, 2014.

\bibitem[Hu et al.(2018)]{8509486}
S.~Hu, L.~Zou, J.~Yu, H.~Wang, and D.~Zhao.
\newblock Answering natural language questions by subgraph matching over
  knowledge graphs (extended abstract).
\newblock In {\em 2018 IEEE 34th International Conference on Data Engineering
  (ICDE)}, pages 1815--1816, Los Alamitos, CA, USA, apr 2018. IEEE Computer
  Society.

\bibitem[Li et al.(2009)]{4633358}
Juanzi Li, Jie Tang, Yi~Li, and Qiong Luo.
\newblock Rimom: A dynamic multistrategy ontology alignment framework.
\newblock {\em IEEE Transactions on Knowledge and Data Engineering},
  21(8):1218--1232, 2009.

\bibitem[Manning et al.(2014)]{manning-EtAl:2014:P14-5}
Christopher~D. Manning, Mihai Surdeanu, John Bauer, Jenny Finkel, Steven~J.
  Bethard, and David McClosky.
\newblock The {Stanford} {CoreNLP} natural language processing toolkit.
\newblock In {\em Association for Computational Linguistics (ACL) System
  Demonstrations}, pages 55--60, 2014.

\end{thebibliography}

\section*{Checklist}


\begin{enumerate}

\item For all authors...
\begin{enumerate}
  \item Do the main claims made in the abstract and introduction accurately reflect the paper's contributions and scope?
    \answerYes{}
  \item Did you describe the limitations of your work?
    \answerYes{}
  \item Did you discuss any potential negative societal impacts of your work?
    \answerYes{}
  \item Have you read the ethics review guidelines and ensured that your paper conforms to them?
    \answerYes{}
\end{enumerate}

\item If you are including theoretical results...
\begin{enumerate}
  \item Did you state the full set of assumptions of all theoretical results?
    \answerYes{}
	\item Did you include complete proofs of all theoretical results?
    \answerYes{}
\end{enumerate}

\item If you ran experiments...
\begin{enumerate}
  \item Did you include the code, data, and instructions needed to reproduce the main experimental results (either in the supplemental material or as a URL)?
    \answerYes{}
  \item Did you specify all the training details (e.g., data splits, hyperparameters, how they were chosen)?
    \answerYes{}
	\item Did you report error bars (e.g., with respect to the random seed after running experiments multiple times)?
    \answerYes{}
	\item Did you include the total amount of compute and the type of resources used (e.g., type of GPUs, internal cluster, or cloud provider)?
    \answerYes{}
\end{enumerate}

\item If you are using existing assets (e.g., code, data, models) or curating/releasing new assets...
\begin{enumerate}
  \item If your work uses existing assets, did you cite the creators?
    \answerYes{}
  \item Did you mention the license of the assets?
    \answerYes{}
  \item Did you include any new assets either in the supplemental material or as a URL?
    \answerYes{}
  \item Did you discuss whether and how consent was obtained from people whose data you're using/curating?
    \answerYes{}
  \item Did you discuss whether the data you are using/curating contains personally identifiable information or offensive content?
    \answerYes{}
\end{enumerate}

\item If you used crowdsourcing or conducted research with human subjects...
\begin{enumerate}
  \item Did you include the full text of instructions given to participants and screenshots, if applicable?
    \answerNA{}
  \item Did you describe any potential participant risks, with links to Institutional Review Board (IRB) approvals, if applicable?
    \answerNA{}
  \item Did you include the estimated hourly wage paid to participants and the total amount spent on participant compensation?
    \answerNA{}
\end{enumerate}

\end{enumerate}

\end{document}